\documentclass[10pt,twocolumn,letterpaper]{article}

\usepackage{cvpr}              % To produce the CAMERA-READY version
\usepackage{wrapfig}
\usepackage{makecell} 
\usepackage{multirow}
\usepackage{bbding} 
\usepackage{pifont}
\usepackage{booktabs}
\usepackage{balance}
\usepackage[dvipsnames]{xcolor}

\definecolor{cvprblue}{rgb}{0.21,0.49,0.74}
\usepackage[pagebackref,breaklinks,colorlinks,citecolor=cvprblue]{hyperref}
\usepackage{longtable}
\usepackage{tikz}
\newcommand*\circled[1]{\tikz[baseline=(char.base)]{
            \node[shape=circle,draw,inner sep=0.2pt] (char) {#1};}}

\def\paperID{12855} % *** Enter the Paper ID here
\def\confName{CVPR}
\def\confYear{2024}

\title{LUWA Dataset: Learning Lithic Use-Wear Analysis on Microscopic Images}

\author{Jing Zhang$^{1,}\thanks{Equal contribution}$,  Irving Fang$^{1,}\footnotemark[1]$,  Juexiao Zhang$^{1}$, Hao Wu$^{1}$, Akshat Kaushik$^{1}$, Alice Rodriguez$^{1}$,\\ Hanwen Zhao$^{1}$, Zhuo Zheng$^{2}$, Radu Iovita$^{1}$, Chen Feng$^{1,}\footnotemark[2]$\\
$^{1}$New York University \quad $^{2}$Stanford University\\
{\tt\small \url{https://ai4ce.github.io/LUWA/}}
\vspace{-6mm}
}
\begin{document}
% \maketitle
\twocolumn[{%
\renewcommand\twocolumn[1][]{#1}%
\maketitle
\vspace{-3mm}
\centering
\includegraphics[width=0.95\textwidth]{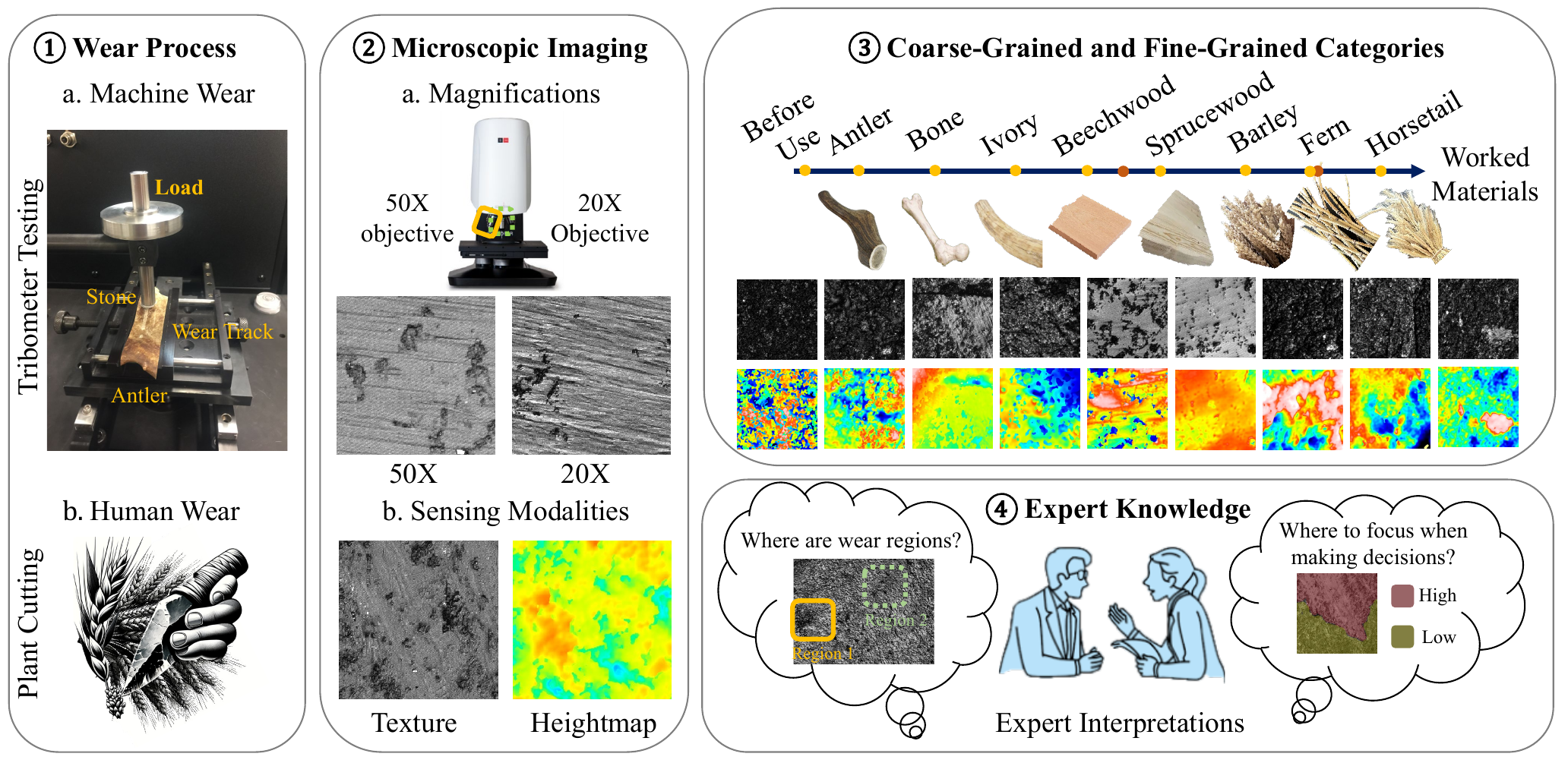}
\captionof{figure}{LUWA poses a unique computer vision challenge due to: \textit{its complex wear formation and irregular wear patterns}, \textit {ambiguous sensing modalities and magnifications in microscopic imaging}. Facing these challenges, the LUWA dataset encompasses both texture and heightmap with different magnifications, encouraging the exploration of image classification beyond common objects.}\vspace{+2mm}
\label{fig:01_fig1}
}]
{
  \renewcommand{\thefootnote}%
    {\fnsymbol{footnote}}
  \footnotetext[1]{Equal contribution}
  \footnotetext[2]{Corresponding author: Chen Feng{\tt\small (cfeng@nyu.edu)}}
}
\begin{abstract}
Lithic Use-Wear Analysis (LUWA) using microscopic images is an underexplored vision-for-science research area. It seeks to distinguish the worked material, which is critical for understanding archaeological artifacts, material interactions, tool functionalities, and dental records. However, this challenging task goes beyond the well-studied image classification problem for common objects. It is affected by many confounders owing to the complex wear mechanism and microscopic imaging, which makes it difficult even for human experts to identify the worked material successfully. In this paper, we investigate the following three questions on this unique vision task for the first time:(\textbf{i}) How well can state-of-the-art pre-trained models (like DINOv2) generalize to the rarely seen domain? (\textbf{ii}) How can few-shot learning be exploited for scarce microscopic images? (\textbf{iii}) How do the ambiguous magnification and sensing modality influence the classification accuracy? To study these, we collaborated with archaeologists and built the first open-source and the largest LUWA dataset containing 23,130 microscopic images with different magnifications and sensing modalities. Extensive experiments show that existing pre-trained models notably outperform human experts but still leave a large gap for improvements. Most importantly, the LUWA dataset provides an underexplored opportunity for vision and learning communities and complements existing image classification problems on common objects.
\vspace{-5mm}
\end{abstract}    
\section{Introduction}
\label{sec:intro}

Lithic Use-Wear Analysis (LUWA) is a long-standing scientific problem (see ~\cref{fig:01_fig1}) to identify the functions of stone tools by examining wear traces at the microscopic level on the tool's surface~\cite{keeley1980experimental, odell1980verifying}. It seeks to distinguish the worked material (like bone, wood, ivory and antler) using microscopic images, creating a classification problem. Investigating this unanswered vision-for-science problem will provide invaluable insights for uncovering the hidden story of ancient tools~\cite{mcpherron2010evidence} and advancing the understanding of material interactions~\cite{teoh2000fatigue, wang2022tooth}. However, few studies have begun to explore advanced learning-based methods, and this field is still in a nascent stage~\cite{byeon2019automated, pizarro2023deep}.
%去掉3 collobrate with experts show exp5 对应的example 
% From the aspect of the application and impects: 1. definition 2. background to demonstrate its impects (using citations form archeaology with the original/most cited/latest literature 3. status with few learning-based methods

Besides its potential scientific impact, learning-based LUWA on microscopic images also poses unique computer vision challenges beyond common objects. Unlike common objects with distinct boundaries and typical size, wear traces in microscopic images are \textit{irregular and discontinuous} (see Fig.~\ref{fig:01_fig1}), making it difficult to define clear visual characteristics~\cite{odell2001stone}. In particular, their absence of clear foreground and background might increase the difficulty of feature extraction. Furthermore, recorded wear traces are \textit{a hybrid consequence of the complex wear process and microscopic imaging}, which is affected by many factors~\cite{odell1980verifying}. For example, the wear duration and the motion type that generates the wear can create significant intra-class variability. Another crucial aspect is \textit{the ambiguous magnification and sensing modality of microscopic images}. High-sensitivity microscopic imaging allows for capturing complementary sensing modalities and zoom-in details at different magnifications, respectively (see Fig.~\ref{fig:01_fig1}). We found that their modality and magnification differences lead to very different visual features. However, due to the understudied characteristics of wear traces, there has been no conclusive answer to this vision-for-science problem about which magnification and sensing modality is better. In practice, the accessibility of expensive equipment also limits the flexible selection of captured data. 
% From the aspect of learning and vision: % 1) pose an unique challenge due to 
%斜体

Recent advances in computer vision provide a good opportunity for this challenging scientific problem, especially the emerging paradigm shift with foundation models pre-trained on large-scale data~\cite{bommasani2021opportunities, awais2023foundational}. These foundation models generate task-agnostic visual features and have shown excellent performance on image classification with common datasets like ImageNet-$\left\{1k,A\right\}$~\cite{oquab2023dinov2}. But how much can state-of-the-art (SOTA) pre-trained representations benefit rarely seen domains not extensively covered by uncurated data available on the Internet? The lack of diversity in domain-specific datasets leaves uncertainty about this question, particularly when addressing real-world vision challenges. Unfortunately, existing studies on LUWA also focus on their own private data and the research community faces a deficiency in accessible datasets. The complexity of wear forms, microscopic imaging, equipment limitations, and the need for expert interpretations hinders the collection of an appropriate dataset that can represent the variability of this domain. 

% 放在后面 machine wear and human wear， 1，3换下
To explore these unique challenges and opportunities, we collaborated with anthropological archaeologists and built the first open-source and the largest LUWA dataset containing 23,130 microscopic images. Its major characteristics are summarized as follows: (\textbf{i}) \textbf{Multi-scale wear patterns.} Lower and higher magnifications allow for an observation of pattern distributions and topographical details, respectively, which increases the scale variations. (\textbf{ii}) \textbf{Complementary sensing modalities.} Both grayscale microscopic images and corresponding 3D surface profiles are available to provide complementary texture information and geometry cues, helping to identify discriminative features~\cite{ferreri2021multi}. (\textbf{iii}) \textbf{Both machine and human wear processes.} Samples of stone tools were collected from both machine and human wear processes. The former is to isolate the effect of worked material by tightly controlling other factors~\cite{marreiros2020controlled} while the latter is to reflect the complexity of real scenes, especially for newly discovered categories without baseline data~\cite{schoville2016new, pedergnana2017monitoring}. We envision that the LUWA dataset will encourage researchers to develop and evaluate algorithms for image classification beyond common objects, as a precursor for downstream tasks like segmentation and detection~\cite{bafghi2023new}. 

Based on the LUWA dataset, we go further in answering the following three problems facing real-world applications: (\textbf{i}) How well representative classification models can generalize to the rarely seen domain? (\textbf{ii}) How can few-shot learning be exploited when scarce microscopic images are available, especially for newly discovered categories? (\textbf{iii}) How do the ambiguous magnification and sensing modality of microscopic images influence the classification accuracy? Our contributions are summarized as follows:
\begin{itemize}
    \item We introduce the first open-source and the largest LUWA dataset including 23,130 microscopic images with different magnifications and sensing modalities. We collaborated with archaeologists and set up a data collection pipeline considering the influence of wear formation, microscopic imaging, and expert knowledge. The LUWA dataset allows for reproducible investigations on this rarely seen domain and complements existing image classification problems on common objects.
    \item Facing the image classification problem beyond common objects, we benchmark the generalization capability of state-of-the-art vision models (ResNet, ViT, DINOv2, ConvNeXt) in this specific domain. Experimental results show that DINOv2 has the most stable performance amidst varying levels of granularity, magnification, and sensing modalities in the data. We also observed many trends regarding the impacts that features like magnification and sensing modality have on classification accuracy, and some of them are not consistent with experts' heuristics. In general, state-of-the-art computer vision models display super-human accuracy over domain experts.
    % Despite its superior performance relative to other methods, visual interpretations show that it regards significant wear traces of some materials (like sprucewood, bone, and antler) as background information.  
    \item Considering the scarcity of microscopic images in real scenarios, we investigate the performance of few-shot image classification and reasoning on the LUWA dataset. Particularly, we collected prompts from archaeologists and did case studies on whether GPT-4V(ision) can mimic the experts' reasoning process. Further explorations are required to improve the performance.
\end{itemize} 
% 3) facing these challlenges what we do
% three contributions: 1) dataset 2) set up a challenge 3) explore 

\section{Related Work}
\label{sec:related}

\noindent\textbf{Lithic Use-Wear Analysis.} Lithic use-wear analysis was originally developed in the 1950s and aims to answer the scientific problem of how to distinguish the worked material according to wear traces on the stone surface~\cite{semenov1964prehistoric, keeley1980experimental, odell1980verifying}. Low-power and High-power microscopy methods provide useful magnified polishes or micro-fractures via confocal microscopes, tactile profilometers, scanning electron microscopes, and even atomic force microscopes~\cite{kimball1995microwear, stemp2001ubm, evans2008laser}. Existing studies focus on blind tests~\cite{newcomer1986investigating, bamforth1988investigating, evans2014importance} and quantification methods~\cite{evans2008laser, stemp2015surface, bartkowiak2019multiscale, bamforth1988investigating, ibanez2019identifying, stevens2010practical}. However, it remains an insufficiently developed research area due to complex wear patterns and the subjectivity of methods~\cite{rodriguez2022effect}. Existing blind tests demonstrate unreliable identification results on different worked materials. For example, correct identifications of plant and wood are 32.4\% and 49.1\%, respectively~\cite{evans2014importance}. The identification of worked materials can be regarded as a unique vision problem, but learning-based algorithms are rarely explored in this research area~\cite{byeon2019automated}. The deficiency in accessible datasets also limits its research progress heavily~\cite{evans2014importance}. To solve these, we present the first open-source LUWA dataset and investigate the capability of SOTA classification algorithms on this unique vision-for-science problem for the first time.
%1.dataset 2. objectivity

\noindent\textbf{Image Classification beyond Common Objects.} Image classification is a fundamental vision task that categorizes a whole image as a specific label or class based on its visual contents. Outstanding performance can be achieved on the image classification task of common objects with representative frameworks like ConvNeXt using only public training data~\cite{woo2023convnext}. However, real-world scenarios, especially scientific applications, involve non-trivial objectives with different physical properties like tiny pollen grains~\cite{battiato2020detection}, constituent materials~\cite{Tanaka_2017_CVPR, Dashpute_2023_CVPR, drehwald2023one}, texture in the wild~\cite{cimpoi2014describing}, and hyperspectral remote sensing images~\cite{scheibenreif2023masked}. Facing uncommon objects, many studies focus on specialized models in respective fields. However, we seek to answer whether simpler pipelines can provide comparable performance. Complementing existing datasets, we introduce a new vision challenge to identify irregular and discontinuous wear traces at the microscopic level and build a classification benchmark to explore how far the practical performance of image classification on uncommon objects can be pushed utilizing existing datasets and architectures, especially foundation models.
%1. definifion 2. success on common objects 3. real-world challenges beyond common objects 4. pipeline 5. dataset 

\noindent\textbf{Microscopic Image Classification.} In the context of using microscopic images for classification, a wide variety of applications can be found across different scientific disciplines. This includes the study of biological cells~\cite{jackson2020single, chen2016deep}, bacteria~\cite{wang2020early}, tissue types~\cite{vu2019methods}, and material structures~\cite{hayman2004significance}. Each application presents unique challenges, particularly in terms of the high-level detail and focus required in the images. These challenges are often compounded by issues such as ambiguity in magnification and sensing modality~\cite{yang2018assessing, hayman2004significance}. To alleviate these challenges, the LUWA Dataset presents a diverse configuration that covers different magnification levels and sensing modalities.
\section{LUWA Dataset}

\begin{figure*}
  \centering
    \includegraphics[width=\linewidth]{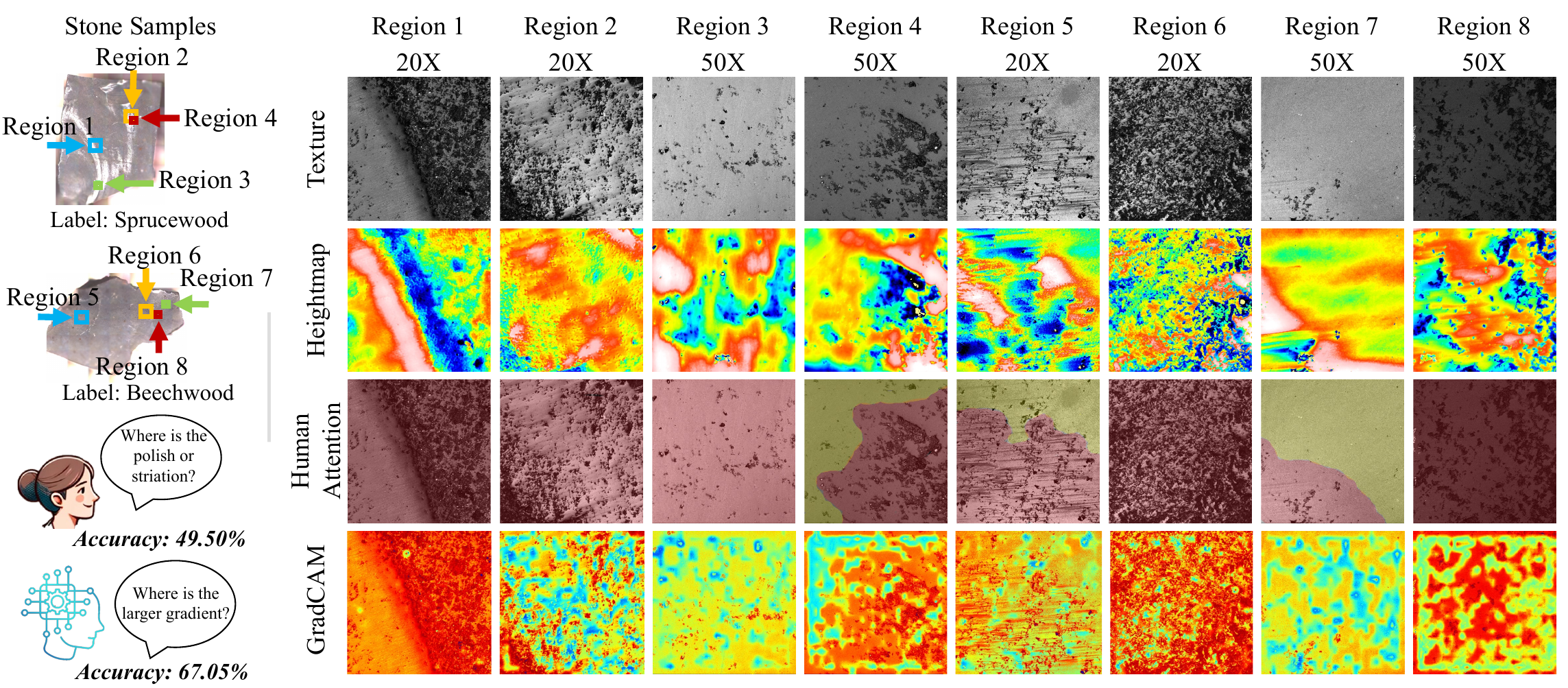}
    \caption{\textbf{Image diversity of LUWA dataset and corresponding visual explanations for human and model decision-making processes.} \textbf{(i)} LUWA dataset provides diverse microscopic images associated with spatial distributions (e.g. Regions 1 and 2), magnifications (e.g. Regions 2 and 4) and sensing modalities (texture in the first row and heightmap in the second row); \textbf{(ii)} We compared visual explanations in both human (in the third row) and model (in the fourth row) decision-making processes. Human experts labeled the most important region with red and the less important region with yellow when looking at details of microscopic images to distinguish the worked material. Similarly, Grad-CAM~\cite{selvaraju2017grad} heatmaps use red for the highest importance, yellow for lower importance, and blue for the lowest importance. Interestingly, similar areas (e.g. Regions 1, 4 and 6) are labeled with higher importance for both humans and models.}
    \label{fig:03_fig1}
    \vspace{-0.5cm}
\end{figure*}

In this section, we describe the LUWA dataset creation process and provide basic statistics. To represent the variability of this domain, in Section~\ref{sec:data_collection}, we present a data collection pipeline, which consists of four key aspects (see ~\cref{fig:01_fig1}). Specifically, \circled{1} considering the complexity of wear formation, we introduced both machine and human wear experiments ~\cite{marreiros2020rethinking} to create stone samples; \circled{2} to enrich the dataset diversity and investigate the ambiguous magnification and sensing modality, we utilized an optical 3D profilometer with both $20\times$ and $50\times$ objective lenses to acquire high-quality texture and heightmap; \circled{3} natural materials were selected according to existing blind tests in the literature~\cite{evans2014importance}, particularly including fine-grained categories of wood and plants. \circled{4} domain-specific knowledge is twofold including the identification of wear degrees to increase the dataset diversity and expert interpretations for potential explorations on explainability and the application of vision language models.  In Section~\ref{sec:data_analysis}, we summarized the LUWA dataset and analyzed its diversity in spatial distributions, magnifications, and sensing modalities. % We split the training and test dataset according to the stone index to ensure robustness against texture biases of the original flint surface;
% overall diagram
% overall

\subsection{Dataset Creation}
\label{sec:data_collection}
%multi-variable  阐述这个名词aim 合适的
\noindent\textbf{Both Machine and Human Wear Processes.} LUWA dataset contains stone samples from both machine and human wear processes. Key factors that affect wear results are material properties, mechanical factors, and environmental conditions~\cite{rymuza2007tribology}. To isolate the effect of worked materials, a tightly controlled protocol was used for machine wear experiments~\cite{marreiros2020rethinking, marreiros2020controlled}. We utilized a tribometer to simulate cutting actions so as to quantify the load applied to the material (a load of 20N), the type of movement (a straight back and forth motion with the speed of 35 repetitions per minute), and the worked duration (0h, 1h, 3h, 5h, and 12h) (see Fig.~\ref{fig:01_fig1}). To limit the influence of material properties of stone samples, we chose the same flint (Baltic/morainic flint from Denmark) for all experiments~\cite{rodriguez2022effect, schmidt2020mineralogy}. Considering the low classification accuracy of various plants in blind tests~\cite{evans2014importance}, we chose the plant cutting process as the human wear experiment. %Furthermore, a practitioner used nine stone tools to cut three types of plants with different silicon contents for nine hours. 
%The variation of the worked duration is to include diverse visual features during the wear formation. All experiments are conducted in similar environmental conditions.  
%For machine wear experiments, we used a low speed precision cutter to handle the flint and potential worked materials and adjusted their size to fit the clamp of the tribometer. We used the natural flat surface of the flint for polishing to restrict the influence of its irregular surface topography.

\noindent\textbf{High-Quality Microscopic Imaging.}
To capture the high-precision wear traces on stone samples, we utilized an optical 3D profilometer (S neox, Sensorfar Metrology) to collect data with a standardized and reproducible process (see Fig.~\ref{fig:01_fig1}). To test the influence of magnifications for microscopic image classification, both $20\times$ and $50\times$ objective lenses were chosen for measurements. Their spatial resolutions are 0.65 and 0.26 $\mu$m/pixel, respectively. Furthermore, complementary grayscale images and corresponding 3D surface profiles are acquired via Sensormap. We applied a standard filtering protocol to extract the worn surface and alleviate the effect of natural flints' surface topography~\cite{calandra2019back}. 
% wear selection, accuracy, magnification, resolution, sensing modality
%Human experts provided assistance to find traces with different degrees of wear to increase the data diversity. 

\vspace{-0.5mm}
\noindent\textbf{Domain-Specific Expert Knowledge.}
Domain-specific knowledge is twofold: (\textbf{i}) human experts help to identify microscopic traces with different wear degrees, enriching the dataset diversity; (\textbf{ii}) for further investigations on explainability and the application of vision language models, human experts also labeled their attention maps when making decisions on worked material (see Fig~\ref{fig:01_fig1}) and provided classification prompt for GPT-4V~\cite{openai2023chatgpt} (see \cref{fig:04_llm_fs}).
% \noindent\textbf{Dataset Division.}
% The uneven distribution of the original flint surface also introduces the challenge for the identification of worked materials. Considering Tobler’s First Law, i.e., everything is related to
% everything else, but near things are more related than distant things~\cite{tobler1970computer, wang2021loveda}, we divided the training and test dataset based on the stone index to make sure that not including microscopic images from the same flint sample into both training and test parts. 

\vspace{-0.5mm}
\noindent\textbf{Material Selection and Processing.}
To benchmark models in this field, we chose representative natural materials (see ~\cref{fig:01_fig1}) according to blind test results in the literature~\cite{evans2014importance} and included fine-grained categories on wood and plants in particular. For the further exploration of wear mechanisms, we analyzed their properties, including the hardness~\cite{rodriguez2022effect} and silicon content~\cite{fullagar1991role} in the supplementary material. 
% natural materials; flint type & shapes
%(antler, bone, ivory, two types of wood for variable-controlling experiments, and three types of plant for nonintervention experiments) according to blind tests 

\begin{table}[]
\begin{center}
\resizebox{\columnwidth}{!}{
\begin{tabular}{cccccccccccccc}
\Xhline{4\arrayrulewidth}
\hline
\textbf{Stone Samples} & \textbf{Motion Types} & \textbf{Worked Time}  & \textbf{Material Categories} \\ \hline
34                     & 2                     & 7                     & \multicolumn{1}{c}{9}   \\ \hline
\multicolumn{2}{c}{\textbf{Magnifications}}    & \multicolumn{2}{c}{\textbf{Sensing Modalities}} \\ \hline
\multicolumn{2}{c}{2}                          & \multicolumn{2}{c}{2}                           \\ \hline
\Xhline{4\arrayrulewidth}
\end{tabular}}
\end{center}
\vspace{-6mm}
\caption{Key factors considered in the LUWA dataset that can reflect the complex wear formation and microscopic imaging.}
\label{table:03_dataset}
\vspace{-6mm}
\end{table}

\subsection{Dataset Analysis}
\label{sec:data_analysis}

% \begin{table*}
%   \centering
%   \begin{tabular}{ccccccc}
% \hline
% \multirow{2}{*}{Images} & \multicolumn{4}{c}{Use Wear}                           & \multicolumn{2}{c}{Microscopic Imaging} \\ \cline{2-7} 
%                         & Stone Samples & Motion Types & Worked Time & Material Classes & Magnifications   & Sensing Modalities   \\ \hline
% 23130                   & 34     & 2            & 7           & 9                & 2                & 2                    \\ \hline
% \end{tabular}
% \caption{
% Key factors considered in LUWA dataset that can reflect the complex wear formation and microscopic imaging.
% }
% \vspace{-0.05in}
% \label{tab:03_dataset}
% \end{table*}

% \begin{figure*}
%   \centering
%     \includegraphics[width=\linewidth]{figs/03_fig2.pdf}
%     \caption{Left:example of .Right.}
%     \label{fig:03_fig2}
% \end{figure*}

% \noindent\textbf{Dataset Characteristics.}
To reflect the variability of this scientific domain, we built the first public and largest LUWA dataset containing 23,130 microscopic images. Specifically, key factors of the complex wear formation and microscopic imaging are considered in the LUWA dataset. As shown in ~\cref{table:03_dataset}, we report (\textbf{i}) the number of microscopic images, (\textbf{ii}) the number of stone samples, motion types, worked time, and material classes which are exploited, (\textbf{iii}) the number of magnifications and sensing modalities LUWA dataset supports.

\begin{figure}
  \centering
    \includegraphics[width=\linewidth]{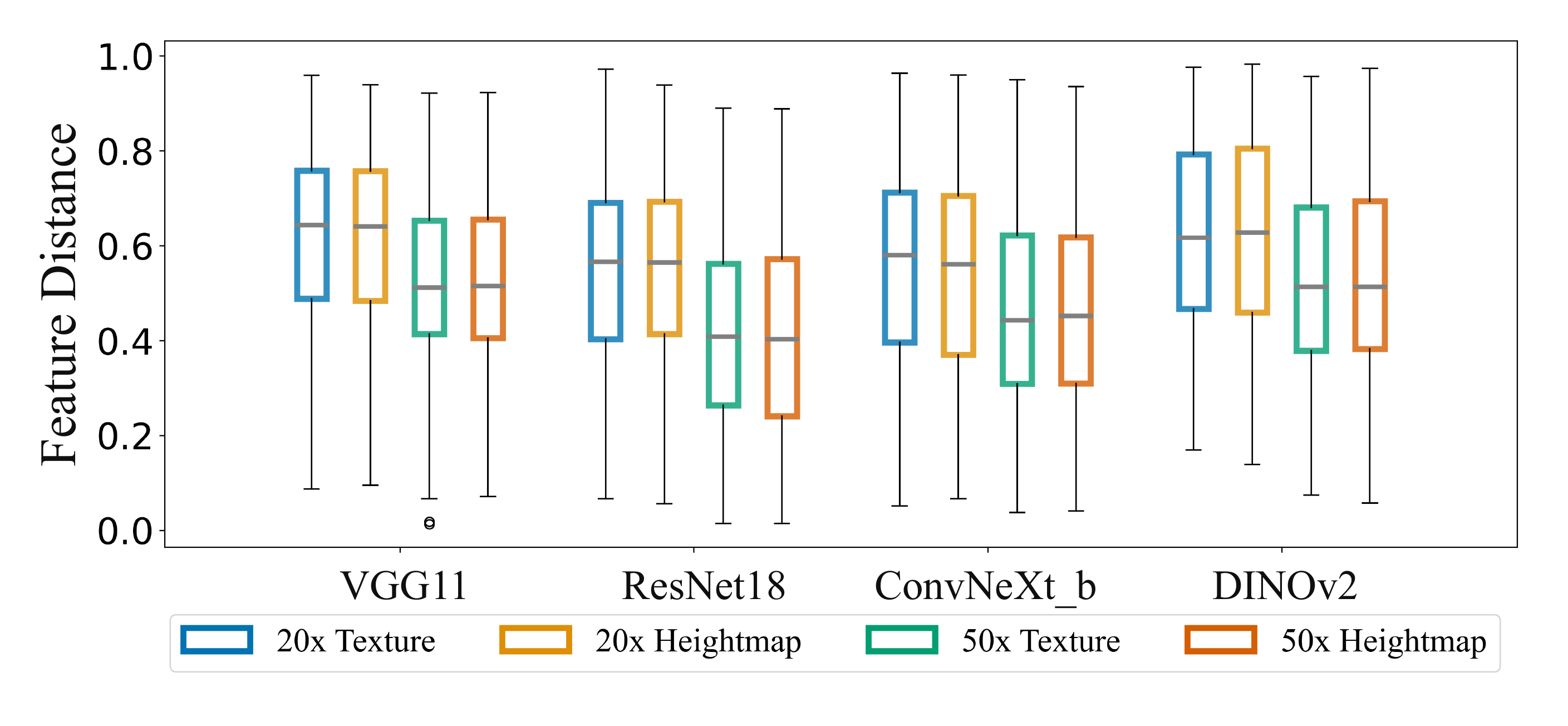}
    \vspace{-0.5cm}
    \caption{Cosine similarity distribution of LUWA dataset on different magnifications and sensing modalities.}
    \vspace{-0.5cm}
    \label{fig:03_fig4}
\end{figure}

Image diversity of LUWA dataset brings challenges for algorithm robustness. It is associated with the spatial distribution of the region collected, the selection of the magnification, and the sensing modality. Greater distances between sampled areas typically result in more pronounced variations in their surface distributions (see Regions 1 and 2 in ~\cref{fig:03_fig1}). Even collected from the same wear trace, the selection of the magnification also contributes to scale difference, which causes totally different wear patterns (see Regions 2 and 4 in ~\cref{fig:03_fig1}). Moreover, LUWA dataset provides both the texture and heightmap, helping to identify discriminative features. We explore the semantic diversity of LUWA dataset on the magnification and sensing modality (see ~\cref{fig:03_fig1})~\cite{bafghi2023new}. We selected VGG~\cite{simonyan2014very}, ResNet~\cite{he2016deep}, ConvNeXt~\cite{liu2022convnet}, and DINOv2~\cite{caron2021emerging, oquab2023dinov2, darcet2023vision} as feature extractors. Then we compute the mean cosine distance of images with different magnifications and sensing modalities, respectively. Scale difference leads to obvious diversity of semantic information and visual descriptions (see ~\cref{fig:03_fig4}).   

% LUWA dataset also supports fine-grained analysis on difficult material categories for image classification. Existing blind tests demonstrate that it is more difficult to distinguish wear marks of plants and wood than ones of antlers and bone. The hardness of wood and the silicon content of plants may influence visual features of polish which are highly reflective with high areas rounded or domed (see Fig.~???????????????\ref{fig:01_fig1}). To investigate these, LUWA dataset supports fine-grained analysis on representative wood (hardness: 0.122 GPa for sprucewood and 2.833 GPa for beechwood) and plants (silicon content: horsetail has the highest silicon content, followed by ferns, and then barley).

% \begin{figure*}
%   \centering
%     \includegraphics[width=\linewidth]{figs/03_fig1.pdf}
%     \caption{Intra-class diversity related to the spatial distribution, magnifications and sensing modalities.}
%     \label{fig:03_fig1}
% \end{figure*}

% \noindent\textbf{Dataset Diversity.}
% \label{sec:data_diversity}

\section{Algorithm Benchmarking}
 By benchmarking a wide range of image classification methods, both classic and state-of-the-art, on this unique vision-for-science dataset, we explore how different features of this dataset affect model performance and hope that we can provide some useful analysis that future work can build on. Specifically, we divide our experiments into two major segments: (1) fully-supervised image classification and (2) few-shot image classification, with specific motivations explained in their corresponding sections below.

\begin{figure*}

  \centering
    \vspace{-0.5cm}
    \includegraphics[width=0.78\linewidth]{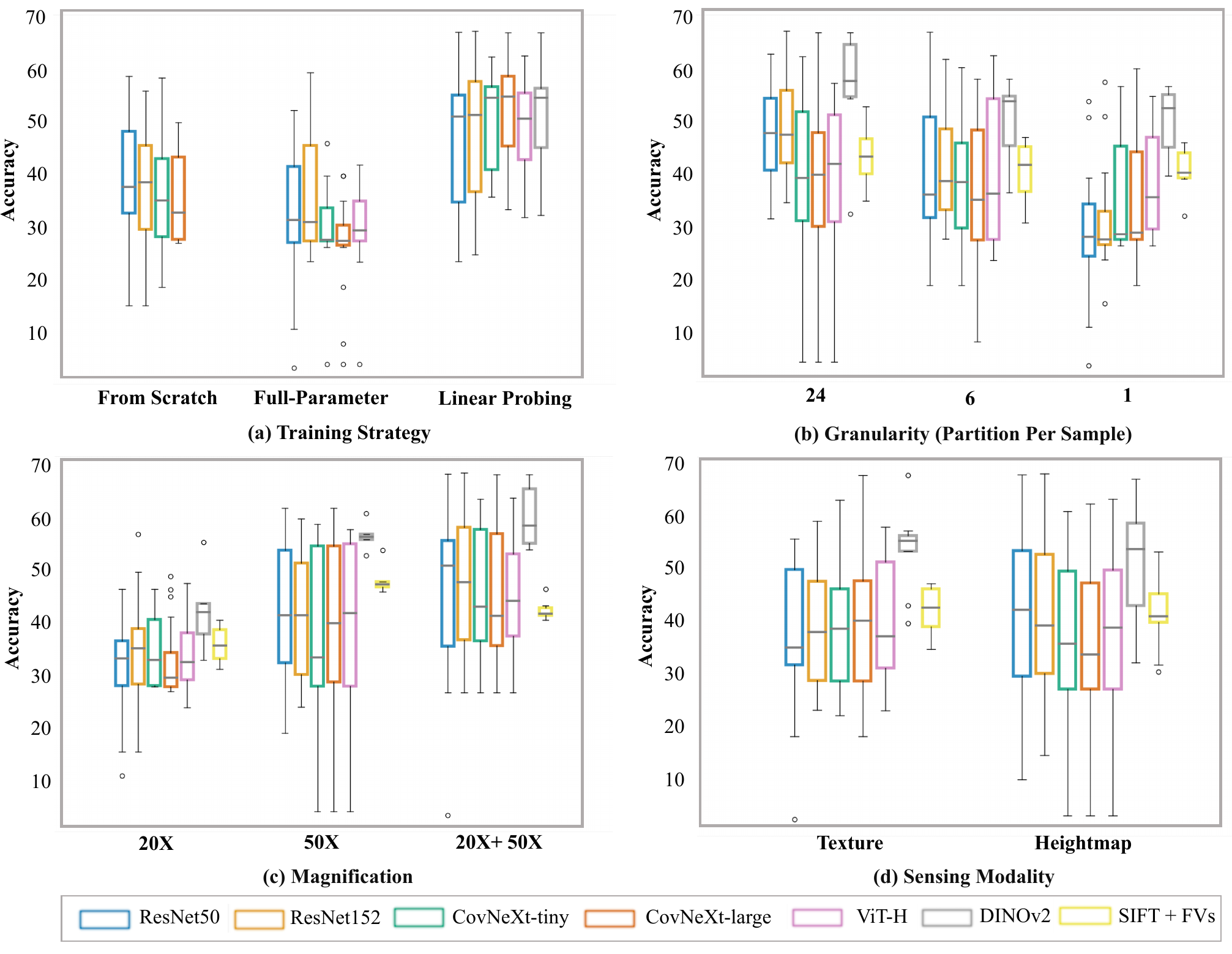}
    \vspace{-0.4cm}
    \caption{The impact of the training strategy, granularity, magnification, and sensing modality on top-1 classification accuracy in \%: (a) Due to their huge parameter counts, the experiments do not include full-parameter fine-tuned DINOv2, and ViT-H and DINOv2 trained from scratch.  (b) Larger numbers in granularity mean more detailed information about a use-wear is fed into the model. }
    \label{fig:04_fig2}
    \vspace{-0.5cm}
\end{figure*}

% \noindent\textbf{Training Details.}Based on the LUWA dataset, we exploit representative algorithms to answer questions tacking real-world challenges: (\textbf{i}) How well representative classification models can generalize to the rarely seen domain? (\textbf{ii}) How few-shot learning can be exploited when scarce microscopic images are available, especially for newly discovered categories? (\textbf{iii}) How the ambiguous magnification and sensing modality of microscopic images influence the classification accuracy?

\subsection{Fully-Supervised Image Classification} 
\label{sec:exp_pre-train}

Unlike datasets crawled from the internet, such as ImageNet-1k \cite{deng2009imagenet} or LAION-5B \cite{LAION-5B}, LUWA contains niche microscopic images with irregular and discontinuous wear traces that often lack obvious foreground or background. In this experiment, we investigate how well classic and state-of-the-art image classification algorithms generalize to the LUWA dataset and seek to find compelling patterns affecting different models' performance. We want to see how well these patterns align with domain experts' knowledge. We also aim to position SOTA methods from the computer vision community with respect to classification performance achieved by human experts. 

\noindent\textbf{Experimental Settings.} We deploy classic methods such as SIFT+FVs \cite{SIFT_FVs}, ResNets\cite{He2015ResNet} and cutting-edge models such as ViT \cite{dosovitskiy2020image}, ConvNeXts \cite{liu2022convnet} and DINOv2 \cite{oquab2023dinov2, darcet2023dino_register} to our LUWA dataset. We believe the time-tested classic methods can serve as a lower-bound benchmark, while the more recent advancements such as DINOv2, which can often be characterized by intensive scale-up in parameter count, can be used as a reference comparable to human experts' performance on the same task.

Another major reason that propels us to deploy this wide range of models is to see if there are consistent trends across different model architectures and parameter counts and if these trends align with domain experts' knowledge. Specifically, we study the impact of image granularity, magnification, and sensing modality on image classification performance. We also compare different training strategies. 

\textit{Granularity} refers to how many pictures one single use-wear is partitioned into. The use-wear is first captured as an image at $865 \times 865$ resolution. Because many pre-trained models resize the input, which results in pixelated images and loss of fine-grained details that many experts believe are crucial to such a classification task, 
we partition the original image into 24 or 6 patches and feed all the patches to the model. Importantly, to make our results comparable to those of human experts, we adopt a \textit{voting mechanism} during test time. If the majority of the 24 patches are classified as class 1, then all the patches of the same use-wear will be classified as class 1 regardless of their actual test results. We believe this method most resembles how archaeologists perform classification when given an $865 \times 865$ image, as they do not assign a label to each partition.

\textit{Magnification} represents the magnification multiplier on our microscopic imaging equipment. Our dataset comprises images after $20\times$ and $50\times$ magnification. We also mix $20\times $ and $50\times$ data together without any indicator of magnification added to the data to see if mixing magnification will cause any confusion in our models. Note that the magnification multipliers are fixed when the images are taken, and $50\times$ images will not look like $20\times$ images, even if we significantly downsize them.

\textit{Sensing Modality} refers to whether
the picture of the use-wear is stored as texture scans or heightmaps. Texture scans have no depth information, although depth cues are still present. Meanwhile, heightmaps explicitly store the depth information and only the depth information. We want to see if different ways to represent the use-wear will affect computer vision models.

\textit{Training Strategy} is also varied in our experiments. Some models are initialized with state-of-the-art initialization methods \cite{kaiming_initialization} and then trained from scratch. We also apply full-parameter fine-tuning and linear probing \cite{linear-probing} with unfrozen and frozen pre-trained weights, respectively.

By combining the above configurations, we have 324 total experiment results.

\noindent\textbf{Implementation Details.} We deploy ResNet50 (25.6M parameters), ResNet152 (60.4M), ConvNeXt-tiny (28.6M), ConvNeXt-large (197.7M), ViT-H (632M), and DINOv2-ViT-g/14 with registers \cite{darcet2023dino_register} (1.1B). All models are trained with Adam optimizer \cite{Adam} on the default setting in PyTorch. We employ linear warmup with cosine annealing \cite{linearwarmup} as the learning rate scheduler strategy. No data augmentation technique is applied during pre-processing. We defer more details to our supplementary material.

\noindent\textbf{Results and Analysis.} The overall experiment results can be found in \cref{fig:04_fig2}. \cref{fig:04_fig2} (a) demonstrated that linear probing yields the most stable and optimal performance across the broad. These results align with our expectation as fine-tuning on uncommon domain-specific datasets may cause catastrophic forgetting~\cite{kirkpatrick2017overcoming}. On the other hand, the LUWA dataset itself is too small to train generalizable models from scratch. From the perspective of granularity (\cref{fig:04_fig2} (b)), the more granular partition of the images tends to result in better outcomes, although a diminishing marginal return can be observed. This aligns with our speculation that keeping the original larger image’s information as intact as possible is beneficial. It is also possible that the introduction of a voting mechanism brought about a positive regularization effect. More discussion on the voting mechanism can be found in the supplementary material. Considering the selection of magnification, results in \cref{fig:04_fig2} (c) indicated that a higher magnification multiplier is beneficial, which aligns with some human experts' opinions. Notably, mixing data with different magnifications does not confuse the models, and they are able to reap the benefit of abundant data. The same cannot be said for humans, as images with different magnifications can cause confusion. For the sensing modality, we observed that while the best results are usually trained with heightmaps, larger models tend to favor texture. However, the discrepancy is small in general, and the overall performances of the two modalities are comparable as in \cref{fig:04_fig2} (d). More visualizations and detailed tables that follow the trends described above can be found in the supplementary material.

\begin{table*}[]
\begin{center}
\resizebox{2\columnwidth}{!}{
\begin{tabular}{cccccccccccccc}
\Xhline{4\arrayrulewidth}
\hline
\multirow{2}{*}{\textbf{PreTr}} & \multicolumn{2}{c}{\textbf{20X TEX}}                  & \multicolumn{2}{c}{\textbf{50X TEX}}                  & \multicolumn{2}{c}{\textbf{20X HM}}                   & \multicolumn{2}{c}{\textbf{50X HM}}                   & \multicolumn{2}{c}{\textbf{20X+50X TEX}}              & \multicolumn{2}{c}{\textbf{20X+50X HM}}               \\ \cline{2-13} 
                                & \textbf{9w5s}             & \textbf{9w20s}            & \textbf{9w5s}             & \textbf{9w20s}            & \textbf{9w5s}             & \textbf{9w20s}            & \textbf{9w5s}             & \textbf{9w20s}            & \textbf{9w5s}             & \textbf{9w20s}            & \textbf{9w5s}             & \textbf{9w20s}            \\ \hline
ResNet18                        & 54.54                     & 61.97                     & 54.43                     & 62.48                     & 31.19                     & 38.79                     & 35.27                     & 42.34                     & 42.11                     & 49.60                     & 26.43                     & 31.80                     \\
ResNet50                        & 54.13                     & 59.20                     & 55.08                     & 62.18                     & 32.67                     & 38.97                     & 36.71                     & 43.95                     & 45.37                     & 51.46                     & 28.91                     & 34.56                     \\
ResNet152                       & 52.92                     & 59.14                     & 57.59                     & 64.26                     & 30.83                     & 38.74                     & 34.12                     & 41.40                     & 44.32                     & 51.40                     & 26.39                     & 31.89                     \\ \hline
ConvNeXt-tiny                   & 46.27                     & 52.44                     & 52.74                     & 59.23                     & 32.25                     & 39.72                     & 36.43                     & 43.46                     & 42.64                     & 49.43                     & 27.33                     & 33.04                     \\
ConvNeXt-base                   & \multicolumn{1}{l}{48.04} & \multicolumn{1}{l}{54.45} & \multicolumn{1}{l}{54.74} & \multicolumn{1}{l}{62.48} & \multicolumn{1}{l}{31.62} & \multicolumn{1}{l}{39.70} & \multicolumn{1}{l}{35.26} & \multicolumn{1}{l}{43.46} & \multicolumn{1}{l}{41.91} & \multicolumn{1}{l}{48.56} & \multicolumn{1}{l}{26.35} & \multicolumn{1}{l}{32.12} \\
ConvNeXt-large                  & 50.89                     & 57.00                     & 56.65                     & 63.51                     & 30.15                     & 37.70                     & 35.20                     & 42.91                     & 43.80                     & 50.67                     & 25.46                     & 30.79                     \\ \hline
ViT-base                        & 41.00                     & 48.80                     & 43.89                     & 50.99                     & 20.60                     & 24.98                     & 22.68                     & 27.59                     & 35.65                     & 42.31                     & 19.32                     & 22.64                     \\ \hline
\textbf{DINO-small}             & \textbf{58.85}            & \textbf{66.10}            & \textbf{59.50}            & \textbf{67.35}            & \textbf{33.55}            & \textbf{41.27}            & \textbf{42.52}            & \textbf{51.11}            & \textbf{46.94}            & \textbf{53.93}            & \textbf{28.02}            & \textbf{33.49}            \\
\textbf{DINO-base}              & \textbf{57.28}            & \textbf{65.33}            & \textbf{61.39}            & \textbf{69.67}            & \textbf{33.07}            & \textbf{41.83}            & \textbf{42.39}            & \textbf{51.23}            & \textbf{47.52}            & \textbf{55.34}            & \textbf{28.23}            & \textbf{34.00}            \\ \hline
DeiT-small                      & 47.00                     & 55.99                     & 52.08                     & 60.64                     & 29.36                     & 36.70                     & 35.45                     & 44.43                     & 39.93                     & 47.68                     & 26.18                     & 32.14                     \\
DeiT-base                       & 53.70                     & 61.48                     & 55.12                     & 63.81                     & 32.71                     & 40.68                     & 37.67                     & 46.80                     & 44.21                     & 52.57                     & 27.20                     & 34.13                     \\
CLIP-base                       & 42.98                     & 51.30                     & 46.52                     & 55.01                     & 29.75                     & 36.92                     & 36.91                     & 44.51                     & 34.45                     & 41.29                     & 27.81                     & 34.02                     \\ \hline
GPT-4V                      & 37.78                     & -                         & 31.11                     & -                         & 20.00                        & -                         & 20.00                        & -                         & 21.11                         & -                         & 23.33                         & -                         \\ \hline
Human Expert                      & 35.00                     & -                         & 43.75                     & -                         & 20.00                        & -                         & 18.75                        & -                         & 33.33                         & -                         & 19.44                         & -                         \\ \hline
\Xhline{4\arrayrulewidth}
\end{tabular}}
\end{center}
\vspace{-0.5cm}
\caption{Few-shot image classification performance on LUWA dataset is associated with the magnification and sensing modality. `PreTr' denotes pre-trained models we used; `20X' and `50X' denote microscopic images at $20\times$ and $50\times$ magnifications; `TEX' and `HM' denote texture and heightmap; `9w5s' and `9w20s' denote 9-way-5-shot and 9-way-20-shot,  respectively.}
\label{table:few-shot-learning}
\vspace{-0.3cm}
\end{table*}

The best performance of $67.05\%$ top-1 accuracy is achieved by linear probing on ImageNet-1K pre-trained ResNet152 with the heightmap data of 24 partitions and $20\times$+$50 \times$ magnification. Overall,
DINOv2 excels across all aspects, demonstrating the most stable performance amidst varying levels of granularity, magnification, and sensing modalities. The worst performance of the traditional baseline SIFT+FVs is better than the worst configurations of other deep learning methods, but its better configurations are significantly worse than those of the deep learning methods that achieved upper-echelon performance.

\noindent\textbf{Explainability and Comparison with Human Experts}. Currently, archaeologists have about $49.5\%$ accuracy in a double-blind test with a similar setup \cite{evans2014importance}. Our in-house testing with two professional archaeologists yields an accuracy of $43.75\%$ in a few-shot setting (\cref{table:few-shot-learning}). All tested models, except SIFT+FVs, are able to achieve far better accuracy (over $59.5\%$) under several configurations of the dataset. Notably, DINOv2 with linear probing is able to achieve superior or comparable performance under almost all possible configurations.

\begin{figure*}
  \centering
    \includegraphics[width=\linewidth]{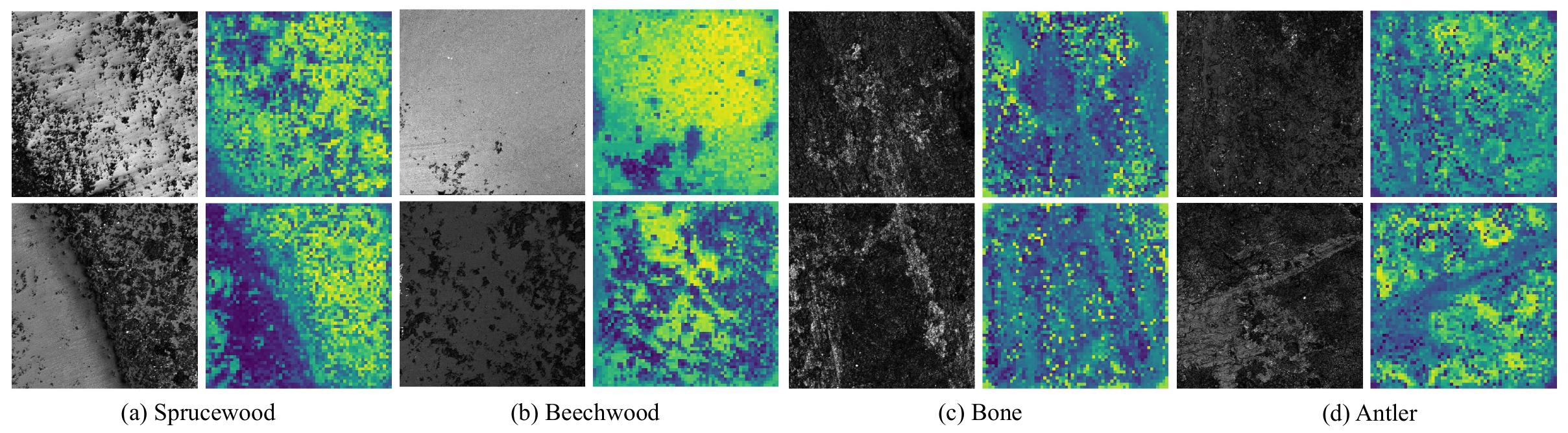}
    \vspace{-0.6cm}
    \caption{Feature visualization of LUWA dataset using frozen pre-trained DINOv2.}
    \label{fig:04_dino_vis}
    \vspace{-0.5cm}
\end{figure*}

However, feature visualization demonstrated that DINOv2 (with registers for better visualization) sometimes recognizes important polished regions in microscopic images of beechwood (see \cref{fig:04_dino_vis}) as the foreground, but recognizes the same polish of different categories ( sprucewood, bone, and antler, see \cref{fig:04_dino_vis}) as the background. More explorations are needed to explain this unwanted behavior. Interestingly, we found that the regions recognized as highly important for classification are similar for human experts and our best model ResNet152, under the same data configuration (see Regions 1, 4, and 6 in \cref{fig:03_fig1}). We visualize the results for ResNet152 using Grad-CAM \cite{selvaraju2017grad}.

% Please add the following required packages to your document preamble:
% \usepackage{multirow}
% Please add the following required packages to your document preamble:
% \usepackage{multirow}

\subsection{Few-Shot Image Classfication}
\label{sec:exp_fsl}

In practice, LUWA faces a scarcity of microscopic images due to limited stone tools, expertise requirements, and specialized equipment, especially when discovering new categories. Human experts can identify new classes of wear traces with a few examples. To emulate this, we investigate whether few-shot learning can be utilized and how microscopic image magnifications and sensing modalities influence the model's performance.

\begin{figure*}[t]
  \centering
    \includegraphics[width=\linewidth]{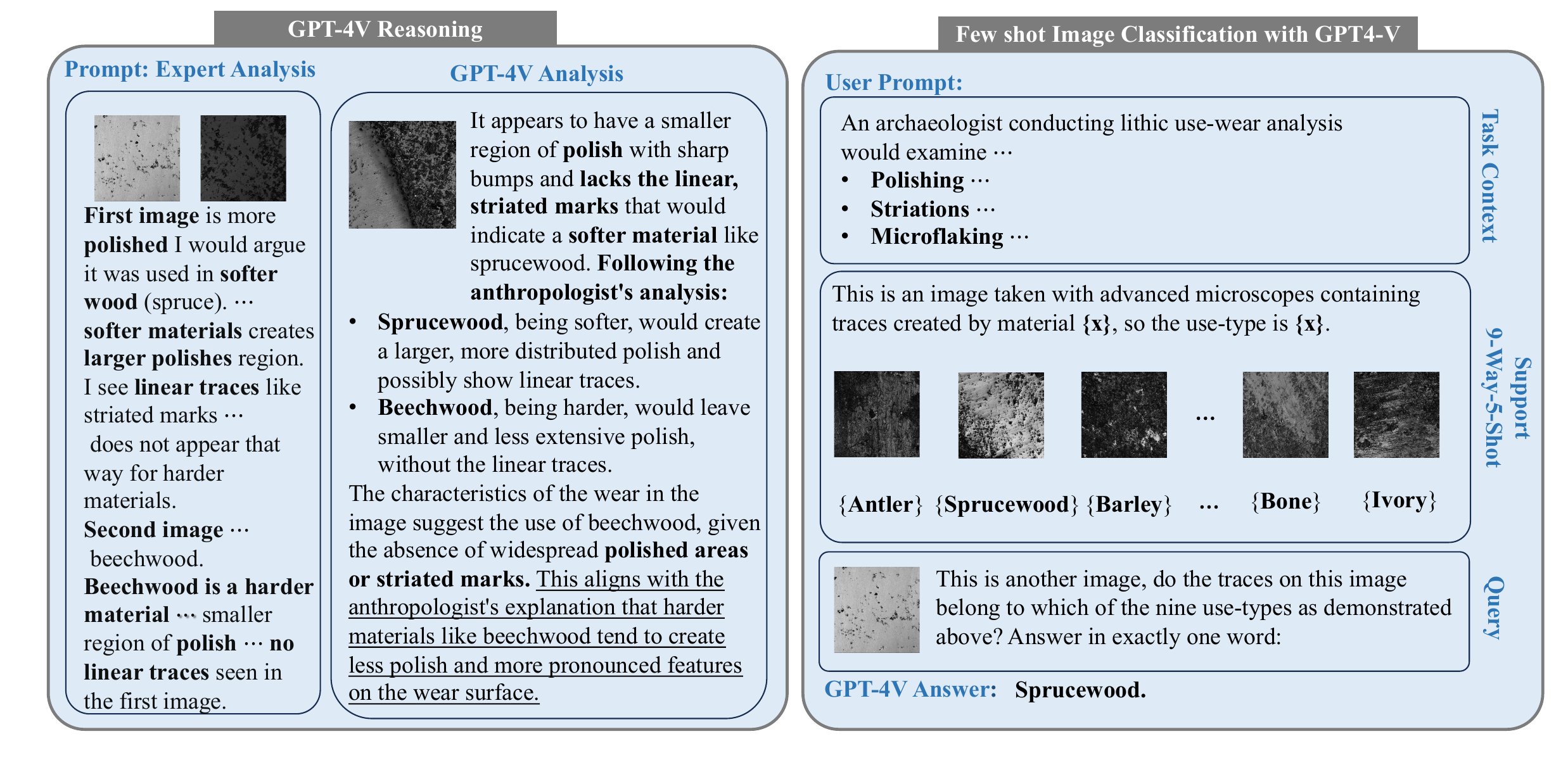}
    \vspace{-0.8cm}
    \caption{GPT-4V few-shot image classification and reasoning following instructions from human experts.}
    \label{fig:04_llm_fs}
    \vspace{-0.5cm}
\end{figure*}

\noindent\textbf{Experimental Settings.} We designed two main experiments: \textbf{(i)} Few-shot image classification with a simple but effective pre-train + ProtoNet pipeline~\cite{hu2022pushing}. We evaluate the performance of powerful pre-trained models (including ResNet, ViT, DINO~\cite{caron2021emerging}, ConvNeXt,  CLIP~\cite{radford2021learning}, DEIT~\cite{touvron2021training}) and popular meta-learners ProtoNet~\cite{snell2017prototypical} on LUWA dataset. We simulated 600 episodes/tasks and results are demonstrated under 9-way-5/20-shot settings. \textbf{(ii)} GPT-4V~\cite{openai2023chatgpt} experiments: few-shot image classification and reasoning following instructions from human experts. To explore the potential mode of AI-human collaboration facing scientific domains in the advent of large multi-modal models, we collected prompts from three archaeologists and conducted case studies on whether the latest GPT-4V can follow and mimic the experts' reasoning process when analyzing the samples. Then we summarized key points that matter during the experts' analysis and used that to prompt the GPT-4V for few-shot image classification and reasoning. The experiments are illustrated in ~\cref{fig:04_llm_fs}. Additionally, we included human experts' test results to reflect the difficulty for humans to distinguish these discovered categories with just a few examples. Results are reported under 9-way-5-shot settings in ~\cref{table:few-shot-learning}.
%both microscopic image numbers and natural worked material categories

% \vspace{-1mm}
\noindent\textbf{Results and Analysis.} Experimental results in \cref{table:few-shot-learning} demonstrated that DINO excels at few-shot learning classification. Note that in the case of limited microscopic images, classification results of textures at $50\times$ magnification setting yield notably superior results compared to others, which provides valuable guidance for few-shot learning tasks in our domain. Moreover, the number of parameters in pre-trained models has a limited impact on this few-shot learning task. We found that GPT-4V can effectively follow the experts' analysis as highlighted in ~\cref{fig:04_llm_fs}. It learns to emphasize the same points that the experts pay attention to. However, the analysis doesn't always lead to the correct answer.
GPT-4V did poorly on few-shot classification. We hypothesize that our data is very different from the web data that GPT models are trained on, and the vision module in GPT-4V still struggles to efficiently present detailed vision information to the language module, especially in a long context such as our multi-image few-shot classification scenario. This means the vision ability of SOTA multi-modal language models still needs improvement before they can be used in scientific tasks with domain-specific data like ours.
% From results of Table~\cref{table:few-shot-learning} we can draw following conclusions: . As illustrated in ~\cref{fig:04_llm_fs}, we prompt GPT-4 Vision to do few-shot learning.

% \jing{analysis of chatgpt}

% Please add the following required packages to your document preamble:
% \usepackage{multirow}

% Please add the following required packages to your document preamble:
% \usepackage{multirow}

% \begin{figure}
%   \centering
%     \includegraphics[width=\linewidth]{figs/04_fig_tsne.pdf}
%     \caption{t-SNE comparison on ground truth and predicted label.}
%     \label{fig:04_fig_tsne}
% \end{figure}

\vspace{-0.2cm}
\section{Impact and Limitations of LUWA Dataset}
\vspace{-0.2cm}
\label{sec:impact}

\noindent\textbf{Scientific Impacts.} AI-expert collaboration can provide invaluable insights for scientific research. To tackle the long-standing problem of stone tool use, we make the first attempt to collaborate with archaeologists and utilize advanced learning-based methods for worked material inference. The LUWA dataset allows for further investigations to advance our understanding of ancient tool use and material processing techniques.

\noindent\textbf{Limitations and Future Directions.} We will enrich the LUWA dataset from the following three aspects: \textbf{(i)} supplement microscopic images with both worn and unworn regions using lower-magnification objectives, allowing wear trace segmentation and detection tasks; \textbf{(ii)} collect images including wear traces caused by different worked material; \textbf{(iii)} increase categories of man-made materials for comprehensive analysis on wear features and material properties.

\section{Conclusion}
\vspace{-0.35cm}
\label{sec:conclusion}
We collaborate with anthropological archaeologists and present the first public and the largest Lithic Use-Wear Analysis (LUWA) dataset benefiting both vision and science domains. The LUWA dataset serves as a benchmark to evaluate the generalization capabilities of advanced models on image classification tasks beyond common objects. Addressing specific challenges of wear formation and microscopic imaging, LUWA offers vital guidance on selecting suitable magnifications and sensing modalities facing different scenarios. Our analysis reveals that SOTA models encounter distinct difficulties when facing these specific challenges. Despite DINOv2's superior performance relative to other methods, it overlooks visual features that archaeologists identify as indicative of wear. We anticipate that the LUWA dataset will stimulate further research into enhancing the adaptability of large-scale models to specialized domains within computer vision.

\small{
% \section*{Acknowledgment}
\textbf{Acknowledgment}. This work is supported by NSF Grant 2152565, and by NYU IT High-Performance Computing
resources, services, and staff expertise. We gratefully acknowledge Sara Borsodi, Felix Devis Kisena, Kat Liu, Eugenia Ochoa, Vita Jackman Kuwabara, Alice Jiang, Meiyu Zhang, and Sriram Koushik for their valuable assistance in collecting the microscopic images.
}
{\small\bibliographystyle{ieeenat_fullname}
\bibliography{main}
}

% WARNING: do not forget to delete the supplementary pages from your submission 
\clearpage
\setcounter{page}{1}
\maketitlesupplementary

\section*{Appendix}
\label{sec:appendix}
\renewcommand{\thesection}{\Alph{section}}
\renewcommand{\thefigure}{\Roman{figure}}
\renewcommand{\thetable}{\Roman{table}}

\setcounter{section}{0}
\setcounter{figure}{0}
\setcounter{table}{0}

This document supplements the main paper as follows:

\renewcommand{\labelenumii}{\Roman{enumii}}

\begin{enumerate}
\item Describe dataset fidelity, material properties and human annotations (supplement Section~\ref{sec:data_collection}).
\item More details about the training recipe and reproducibility (supplement section~\ref{sec:exp_pre-train}).
\item More visualizations and detailed tables (supplement section~\ref{sec:exp_pre-train}).
\item More details about the human expert tests(supplement section~\ref{sec:exp_fsl}).
 \end{enumerate}
\section{LUWA Dataset}

\subsection{Dataset Fidelity}

\noindent\textbf{Archaeological samples.} Archaeologists struggle to reach a consensus on how to identify the worked material on ancient lithic tools because of a lack of ground truth information. LUWA aims to be the first step to building the benchmark and tool that can help archaeologists make more informed decisions as archaeologists believe the underlying physics should remain the same across real-world and lab-made use wear, and models that can work well on lab-made data could be an ancillary input to archaeologists' heuristics. 

\noindent\textbf{Worked time.} We followed a tightly controlled protocol and ``worked time" to reflect various wear degrees. 

\noindent\textbf{Impact of aging and conservation status.} This is minimized because post-depositional alterations are usually visible under the microscope, and archaeologists can exclude pieces with signs of weathering.

\subsection{Material Properties}

Existing studies have indicated that both the hardness of materials and their silicon content can have an impact on the visual features of wear traces. This suggests that the properties of materials being worked or worn play a significant role in shaping the wear patterns observed. In machine wear experiments, we listed the hardness of worked materials for further explorations of wear mechanisms (see~\cref{table:hardness}). In human wear experiments, LUWA dataset supports fine-grained analysis on representative plants: horsetail has the highest silicon content, followed by ferns, and then barley.% of its dryweight being silica content. In some cases, up to 25% of its dry weight can be silica, making it theplant with the highest concentration of this mineral (silicon content: horsetail has the highest silicon content, followed by ferns, and then barley). in plants (silicon content: horsetail has the highest silicon content, followed by ferns, and then barley).

\subsection{Human Annotations}

\begin{figure}
  \centering
    \includegraphics[width=\linewidth]{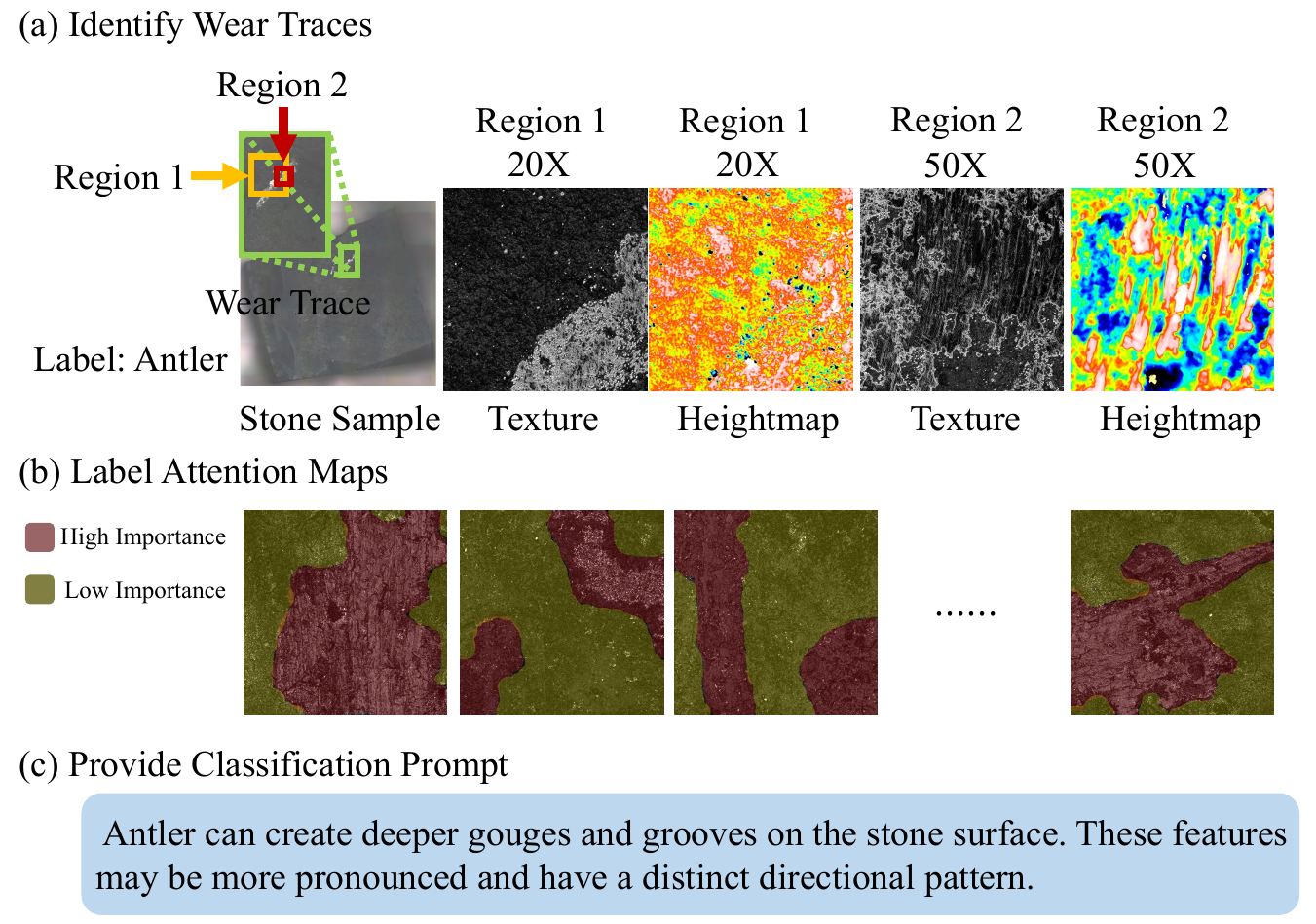}
    \vspace{-0.5cm}
    \caption{Domain-specific expert knowledge: (a) human experts helped to identify wear traces during the process of data collection; (b) human experts labeled the most important region with red and the secondary important region with yellow when making decisions on worked materials; (c) human experts provided classification prompt for GPT-4V.}
    \vspace{-0.5cm}
    \label{fig:a_human}
\end{figure}

Human experts provide domain-specific knowledge for LUWA dataset in the following aspects (see~\cref{fig:a_human}):
\begin{itemize}
    \item Identification of Wear Traces: Human experts are actively involved in the process of data collection and are responsible for identifying wear traces on objects. Their expertise allows them to recognize and differentiate between various types of wear patterns, such as microwear polish, scratches, and impact marks. This identification is fundamental for understanding the history and use of the objects.
    \item Color Labeling for Attention Maps: During the decision-making process regarding worked materials, human experts use a color-coded system to label different regions of the objects. The most important regions are labeled with the color red, while less important regions are labeled with the color yellow. This color-coded system likely helps prioritize the analysis of wear traces and their significance in understanding the function and use of the objects.
    \item Classification Prompt for GPT-4V: Human experts also contribute by providing a classification prompt for GPT-4V, an AI model. This classification prompt likely guides the AI in recognizing and categorizing wear traces on objects, benefiting from the expertise of human specialists to enhance the accuracy of the AI's analysis.
\end{itemize}

% \begin{table}[]
% \begin{center}
% \resizebox{\columnwidth}{!}{
% \begin{tabular}{cccccc}
% \Xhline{4\arrayrulewidth}
% \hline
% \textbf{}                                                                  & \textbf{Ivory}  & \textbf{Antler} & \textbf{Bone}   & \textbf{Beechwood} & \textbf{Sprucewood} \\ \hline
% \textbf{Hardness}                                                          & 3.930±\pm0.025 & 3.253±\pm0.727 & 2.961±\pm0.246 & 2.833±\pm1.672    & 0.122±\pm0.004     \\ \hline
% \textbf{\begin{tabular}[c]{@{}c@{}}Elasticity\\ (Literature)\end{tabular}\begin{tabular}[c]{@{}c@{}}Elasticity\\ (Literature)\end{tabular}\begin{tabular}[c]{@{}c@{}}Elasticity\\ (Literature)\end{tabular}\begin{tabular}[c]{@{}c@{}}Elasticity\\ (Literature)\end{tabular}} & 3-20         & 9-16         & 10-30        & 10-16           & 8-11            \\ \hline
% \Xhline{4\arrayrulewidth}
% \end{tabular}}
% \end{center}
% \caption{Hardness of worked materials in machine wear experiments.}
% \label{table:hardness}
% \vspace{-0.3cm}
% \end{table}

\begin{table}[]
\begin{center}
\resizebox{\columnwidth}{!}{
\begin{tabular}{cccccc}
\Xhline{4\arrayrulewidth}
\hline
\textbf{} & \textbf{Ivory}  & \textbf{Antler} & \textbf{Bone}   & \textbf{Beechwood} & \textbf{Sprucewood} \\ \hline
\textbf{Hardness}         & 3.930$\pm$0.025 & 3.253$\pm$0.727 & 2.961$\pm$0.246 & 2.833$\pm$1.672    & 0.122$\pm$0.004     \\ \hline
\Xhline{4\arrayrulewidth}
\end{tabular}}
\end{center}
\caption{Hardness of worked materials in machine wear experiments.}
\label{table:hardness}
\vspace{-0.3cm}
\end{table}

\section{Algorithm Benchmarking}
\subsection{Training Recipe}
The start learning rate, which is also the $\eta_{max}$ in the linear warmup with cosine annealing scheduler, is set to 0.01. The batch size for the smaller models, such as ResNets and ConvNeXts, is set to 200, while for larger models, such as ViT and DINOv2, it's set to 100 to save VRAM. We do not adjust the learning rate based on changing batch size because we believe our learning rate scheduler will offset the changes. When training from scratch, we train for 20 epochs. We reduce that to 10 epochs when fine-tuning and linear probing. The dataset is partitioned into portions of 6/2/2 as train/val/test. To ensure fairness, we put all images from the same stone sample in the same set. We report all results based on the checkpoints with the lowest validation error. All experiments are done on a single Nvidia A100 with 80 GB VRAM. All models are trained three times with different random seeds and PyTorch \texttt{deterministic=True} and \texttt{benchmark=False} to maximize reproducibility. No data augmentation is applied except simple resizing to $224 \times 224$ to match pre-trained models' input dimension.  
\subsection{More Fully-Supervised Image Classification Results}
We present more results that cannot fit into the main text. 

\begin{figure}
  \centering
    \vspace{-0.5cm}
    \includegraphics[width=1\linewidth]{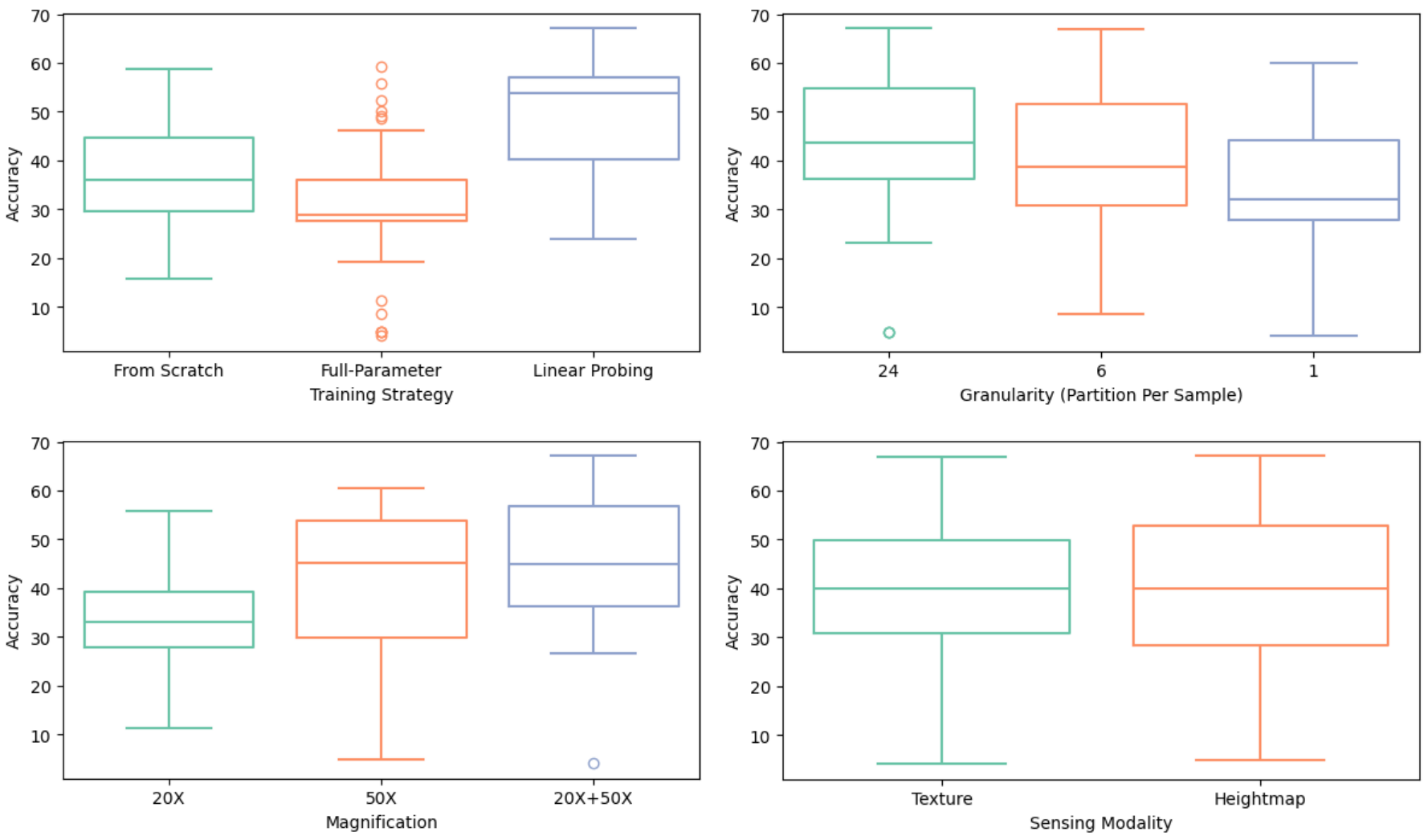}
    \vspace{-0.3cm}
    \caption{The impact of the training strategy, granularity, magnification, and sensing modality on top-1 classification accuracy in \%: Larger numbers in granularity mean more detailed information about a use-wear is fed into the model. }
    \label{fig:suppl_impact}
    \vspace{-0.3cm}
\end{figure}

\begin{figure*}
  \centering
    \includegraphics[width=0.9\linewidth]{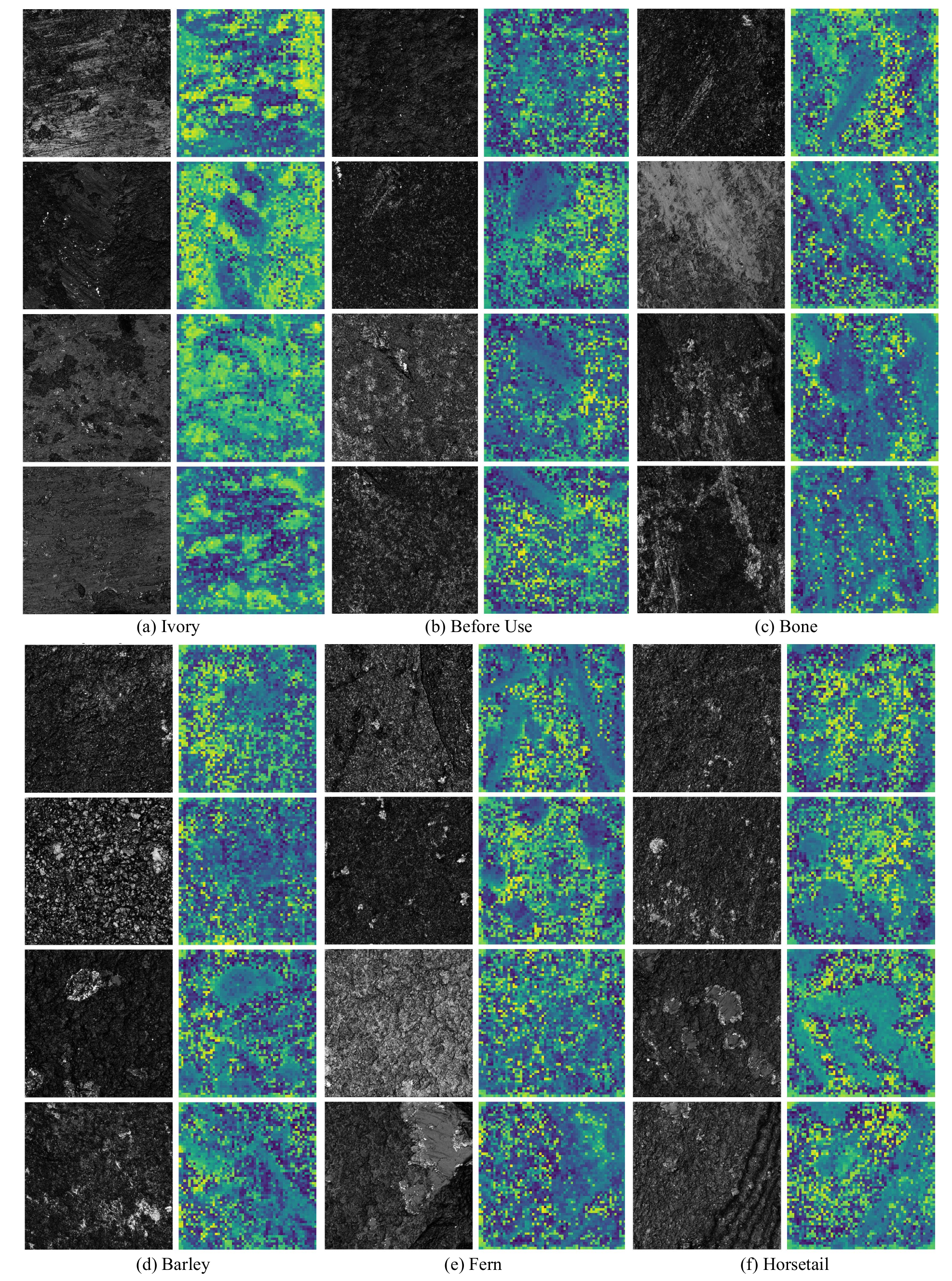}
    % \vspace{-0.5cm}
    \caption{More feature visualization of LUWA dataset using frozen pre-trained DINOv2.}
    \label{fig:suppl_dino_vis}
    \vspace{-0.3cm}
\end{figure*}

\noindent\textbf{More Visualization.} A larger and clearer visualization is contained in \cref{fig:suppl_impact} and \cref{fig:suppl_dino_vis}. As we can see, the trend described in \cref{sec:exp_pre-train} still holds true. 

\noindent\textbf{Quantitative Analysis.} We provide quantitative analysis of the distribution overlap in the regions of interest as described in \cref{sec:exp_pre-train}. As shown in~\cref{tab:IoU}, we select IoU as the quantitative metric for evaluating the distribution overlap.

\begin{table}[]

\end{table}

\begin{table}[h]
	\scriptsize
	\centering
	\vspace{-10pt}
	\caption{IoU for human labeling and Grad-CAM heatmaps.}
	 	\vspace{-8pt}
 	\renewcommand\tabcolsep{1pt}
	\resizebox{1\linewidth}{!}{
		\begin{tabular}{cccccccccc}
\hline
\textbf{}    & \textbf{Fern} & \textbf{Sprucewood} & \textbf{Ivory} & \textbf{Beechwood} & \textbf{Before Use} & \textbf{Horsetail} & \textbf{Barley}                & \textbf{Antler} & \textbf{Bone} \\ \hline
\textbf{IoU} & 0.9089        & 0.8577              & 0.7070         & 0.6959             & 0.6165              & 0.5773             & 0.4929 & 0.4535          & 0.3501        \\ \hline
\end{tabular}}
	\label{tab:IoU}%
 	\vspace{-10pt}
\end{table}%

\begin{table}[]
\begin{center}
\resizebox{\columnwidth}{!}{
\begin{tabular}{cccccc}
\Xhline{4\arrayrulewidth}
\hline
\textbf{Model} & \textbf{Granularity} & \textbf{Magnification} & \textbf{Modality} & \textbf{Training Strategy} & \textbf{Accuracy} \\ \hline
SIFT+FVs & $24$ & $50\times$ & heightmap & N/A  & 52.88 \\ 

ResNet50 & $6$ & $20\times$ + $50\times$ & heightmap & Linear Probing & 66.91 \\ 

ResNet152 & $24$ & $20\times$ + $50\times$ & heightmap & Linear Probing & 67.05 \\ 

ConvNeXt-tiny & $24$ & $20\times$ + $50\times$ & texture & Linear Probing & 62.27 \\ 

ConvNeXt-Large & $24$ & $20\times$ + $50\times$ & texture & Linear Probing & 66.82 \\ 

ViT-H & $6$ & $20\times$ + $50\times$ & heightmap & Linear Probing & 62.5 \\ 

DINOv2 & $24$ & $20\times$ + $50\times$ & texture & Linear Probing & 66.82 \\ 
\hline
\Xhline{4\arrayrulewidth}
\end{tabular}}
\end{center}
\caption{Best Performing Data Configuration for Each Model}
\label{tab:append_best_performing}
\vspace{-0.3cm}
\end{table}

\noindent\textbf{Data Configurations for the Best Performance.} \cref{tab:append_best_performing} shows the data configuration to achieve the best performance for each model. We can see the patterns described in \cref{sec:exp_pre-train} are well reflected among the top-performing models. Note that even though the best model for SIFT+FVs can achieve a reasonable performance of $52.88\%$, most of the other data configurations result in a significant performance downgrade for this method. In fact, this is the only super-human performance ($>49.5\%$ accuracy) for this specific method.

\noindent\textbf{Models that Achieve Super-Human Performance.} \cref{tab:append_superhuman} contains all the models and their corresponding data configurations that achieve super-human performance. Out of 358 possible data configurations, 79 (22\%) are able to achieve super-human performance. \cref{tab:appendix_ratio} contains the count and ratio of different features that appear in super-human models, and we can see that this aligns with the trends described in the main text as well.

\noindent\textbf{More on the Voting Mechanism.}
For the best performing models, \cref{tab:vote} shows that when the final voted prediction is correct, how many partitions are predicted correctly before the voting (Corr Consis), and when the final voted prediction is incorrect, how many partitions are correct (Incorr Consis) or the same as the final wrongly-voted result (Incorr Common Consis). As we can see here, the predictions for each partition are relatively consistent before voting.

\begin{table}[]
\begin{center}
\resizebox{\columnwidth}{!}{
\begin{tabular}{ccc|ccc}
\Xhline{4\arrayrulewidth}
\hline
\textbf{Model Name}    & \textbf{Count} & \textbf{Ratio} & \textbf{Training Strategy} & \textbf{Count} & \textbf{Ratio} \\ \hline
ResNet50               & 16             & 20\%           & Linear Probing             & 66             & 84\%           \\
ResNet152              & 14             & 18\%           & From Sratch                & 8              & 10\%           \\
DINOv2                 & 13             & 16\%           & Full-Parameter Fine-Tuning & 4              & 5\%            \\ \cline{4-6} 
ConvNeXt-tiny          & 13             & 16\%           & \textbf{Granularity}       & \textbf{Count} & \textbf{Ratio} \\
ConvNeXt-large         & 12             & 15\%           & 24                         & 37             & 47\%           \\
ViTH                   & 10             & 13\%           & 6                          & 25             & 32\%           \\
SIFT+FVs               & 1              & 1\%            & 1                          & 17             & 22\%           \\ \hline
\textbf{Magnification} & \textbf{Count} & \textbf{Ratio} & \textbf{Sensing Modality}  & \textbf{Count} & \textbf{Ratio} \\
$20\times$             & 2              & 3\%            & Texture                    & 36             & 46\%           \\
$50\times$             & 38             & 48\%           & Heightmap                  & 43             & 54\%           \\
$20\times$+$50\times$  & 39             & 49\%           & -                          & -              & -              \\ \hline
\Xhline{4\arrayrulewidth}
\end{tabular}}
\end{center}
\caption{Count and ratio of different features that appear in super-human models}
\label{tab:appendix_ratio}
% \vspace{-0.3cm}
\end{table}

\begin{table}[h]
	\scriptsize
	\centering

	 	\
   	\caption{Consistency Analysis of the Voting Mechanism}
 	\renewcommand\tabcolsep{1pt}
	\resizebox{\linewidth}{!}{%
\begin{tabular}{cccc}
\hline
\textbf{Model}    & \textbf{Corr Consis} & \textbf{Incorr Consis} & \textbf{Incorr Common Consis} \\
\hline
\textbf{ResNet50}       & 86.30\%                  & 8.15\%                     & 78.52\%                                \\
\hline
\textbf{ResNet152}      & 78.85\%                  & 11.59\%                    & 62.14\%                                \\
\hline
\textbf{ConvNext-Tiny}  & 82.48\%                  & 12.27\%                    & 60.84\%                                \\
\hline
\textbf{ConvNext-Large} & 78.57\%                  & 9.79\%                     & 66.55\%                                \\
\hline
\textbf{ViT-H}          & 89.80\%                  & 9.33\%                     & 72.00\%                                \\
\hline
\textbf{DINOv2} & 86.34\%  & 7.34\%  & 66.90\% \\
\hline
\end{tabular}%
}
	\label{tab:vote}

\end{table}%

\subsection{More Few-Shot Image Classification Details}
In a test scenario where new categories of wear traces were identified, we provided identical support and query sets to both GPT-4V and two anthropologists. These anthropologists had no prior exposure to the samples in the sets, and we selected their best results for analysis.

\onecolumn
\setlength{\tabcolsep}{4.2mm}{ %set the table width
\setlength{\LTcapwidth}{7in}   %set the table caption width
\begin{longtable}{c|c|c|c|c|c}
\toprule
   Model Name &  Granularity & Magnification &  Sensing Modality &          Training Strategy &  Accuracy \\
\midrule
    ResNet152 &          24 &       $20\times$ + $50\times$ & heightmap &             Linear Probing &     67.05 \\
     ResNet50 &          6 &       $20\times$ + $50\times$ & heightmap &             Linear Probing &     66.91 \\
ConvNeXt-large &          24 &       $20\times$ + $50\times$ &   texture &             Linear Probing &     66.82 \\
       DINOv2 &          24 &       $20\times$ + $50\times$ &   texture &             Linear Probing &     66.82 \\
       DINOv2 &          24 &       $20\times$ + $50\times$ & heightmap &             Linear Probing &     66.14 \\
     ResNet50 &          24 &       $20\times$ + $50\times$ & heightmap &             Linear Probing &     62.73 \\
         ViTH &          6 &       $20\times$ + $50\times$ & heightmap &             Linear Probing &     62.50 \\
 ConvNeXt-tiny &          24 &       $20\times$ + $50\times$ &   texture &             Linear Probing &     62.27 \\
    ResNet152 &          6 &       $20\times$ + $50\times$ & heightmap &             Linear Probing &     61.76 \\
ConvNeXt-large &          24 &       $20\times$ + $50\times$ & heightmap &             Linear Probing &     61.59 \\
     ResNet50 &          24 &           $50\times$ & heightmap &             Linear Probing &     60.58 \\
ConvNeXt-large &          24 &           $50\times$ & heightmap &             Linear Probing &     60.58 \\
 ConvNeXt-tiny &          6 &       $20\times$ + $50\times$ & heightmap &             Linear Probing &     60.25 \\
ConvNeXt-large &          1 &       $20\times$ + $50\times$ & heightmap &             Linear Probing &     60.00 \\
       DINOv2 &          24 &           $50\times$ & heightmap &             Linear Probing &     59.62 \\
    ResNet152 &          24 &       $20\times$ + $50\times$ & heightmap & Full-Parameter Fine-Tuning &     59.32 \\
    ResNet152 &          24 &           $50\times$ & heightmap &             Linear Probing &     58.65 \\
ConvNeXt-large &          1 &           $50\times$ & heightmap &             Linear Probing &     58.65 \\
     ResNet50 &          24 &       $20\times$ + $50\times$ & heightmap &               From Scratch &     58.64 \\
    ResNet152 &          6 &       $20\times$ + $50\times$ &   texture &             Linear Probing &     58.50 \\
 ConvNeXt-tiny &          24 &       $20\times$ + $50\times$ & heightmap &               From Scratch &     58.41 \\
 ConvNeXt-tiny &          6 &       $20\times$ + $50\times$ &   texture &             Linear Probing &     58.09 \\
ConvNeXt-large &          6 &       $20\times$ + $50\times$ & heightmap &             Linear Probing &     58.09 \\
       DINOv2 &          6 &       $20\times$ + $50\times$ & heightmap &             Linear Probing &     58.09 \\
 ConvNeXt-tiny &          6 &           $50\times$ & heightmap &             Linear Probing &     57.69 \\
    ResNet152 &          24 &           $50\times$ &   texture &             Linear Probing &     57.69 \\
ConvNeXt-large &          6 &           $50\times$ & heightmap &             Linear Probing &     57.69 \\
    ResNet152 &          1 &       $20\times$ + $50\times$ & heightmap &             Linear Probing &     57.50 \\
         ViTH &          6 &       $20\times$ + $50\times$ &   texture &             Linear Probing &     57.35 \\
         ViTH &          24 &       $20\times$ + $50\times$ &   texture &             Linear Probing &     57.27 \\
         ViTH &          24 &           $50\times$ & heightmap &             Linear Probing &     56.73 \\
 ConvNeXt-tiny &          24 &           $50\times$ &   texture &             Linear Probing &     56.73 \\
 ConvNeXt-tiny &          1 &       $20\times$ + $50\times$ &   texture &             Linear Probing &     56.67 \\
 ConvNeXt-tiny &          1 &       $20\times$ + $50\times$ & heightmap &             Linear Probing &     56.67 \\
       DINOv2 &          1 &       $20\times$ + $50\times$ &   texture &             Linear Probing &     56.67 \\
    ResNet152 &          24 &       $20\times$ + $50\times$ & heightmap &               From Scratch &     55.91 \\
ConvNeXt-large &          6 &       $20\times$ + $50\times$ &   texture &             Linear Probing &     55.88 \\
    ResNet152 &          24 &           $20\times$ &   texture & Full-Parameter Fine-Tuning &     55.82 \\
       DINOv2 &          1 &           $50\times$ &   texture &             Linear Probing &     55.77 \\
       DINOv2 &          24 &           $50\times$ &   texture &             Linear Probing &     55.77 \\
         ViTH &          6 &           $50\times$ & heightmap &             Linear Probing &     55.77 \\
 ConvNeXt-tiny &          24 &           $50\times$ & heightmap &             Linear Probing &     55.77 \\
     ResNet50 &          6 &           $50\times$ & heightmap &             Linear Probing &     55.77 \\
     ResNet50 &          24 &       $20\times$ + $50\times$ &   texture &             Linear Probing &     55.23 \\
     ResNet50 &          6 &       $20\times$ + $50\times$ & heightmap &               From Scratch &     55.15 \\
    ResNet50 &          24 &           $50\times$ &   texture &             Linear Probing &     54.81 \\
         ViTH &          1 &           $50\times$ & heightmap &             Linear Probing &     54.81 \\
       DINOv2 &          6 &           $50\times$ &   texture &             Linear Probing &     54.81 \\
 ConvNeXt-tiny &          6 &           $50\times$ &   texture &             Linear Probing &     54.81 \\
       DINOv2 &          6 &           $50\times$ & heightmap &             Linear Probing &     54.81 \\
ConvNeXt-large &          24 &           $50\times$ &   texture &             Linear Probing &     54.81 \\
    ResNet152 &          24 &       $20\times$ + $50\times$ &   texture &             Linear Probing &     54.77 \\
 ConvNeXt-tiny &          24 &       $20\times$ + $50\times$ & heightmap &             Linear Probing &     54.55 \\
       DINOv2 &          24 &           $20\times$ &   texture &             Linear Probing &     54.39 \\
    ResNet152 &          6 &           $50\times$ & heightmap &             Linear Probing &     53.85 \\
 ConvNeXt-tiny &          1 &           $50\times$ &   texture &             Linear Probing &     53.85 \\
ConvNeXt-large &          1 &           $50\times$ &   texture &             Linear Probing &     53.85 \\
     ResNet50 &          6 &           $50\times$ &   texture &             Linear Probing &     53.85 \\
         ViTH &          24 &           $50\times$ &   texture &             Linear Probing &     53.85 \\
     ResNet50 &          1 &           $50\times$ & heightmap &             Linear Probing &     53.85 \\
         ViTH &          6 &           $50\times$ &   texture &             Linear Probing &     53.85 \\
       DINOv2 &          1 &       $20\times$ + $50\times$ & heightmap &             Linear Probing &     53.33 \\
     ResNet50 &          24 &       $20\times$ + $50\times$ &   texture &               From Scratch &     53.18 \\
       DINOv2 &          6 &       $20\times$ + $50\times$ &   texture &             Linear Probing &     52.94 \\
          SIFT+FVs &          24 &           $50\times$ & heightmap &                        NaN &     52.88 \\
ConvNeXt-large &          6 &           $50\times$ &   texture &             Linear Probing &     52.88 \\
 ConvNeXt-tiny &          1 &           $50\times$ & heightmap &             Linear Probing &     52.88 \\
     ResNet50 &          24 &       $20\times$ + $50\times$ & heightmap & Full-Parameter Fine-Tuning &     52.27 \\
    ResNet152 &          6 &           $50\times$ &   texture &             Linear Probing &     51.92 \\
       DINOv2 &          1 &           $50\times$ & heightmap &             Linear Probing &     51.92 \\
     ResNet50 &          6 &       $20\times$ + $50\times$ &   texture &             Linear Probing &     51.47 \\
         ViTH &          1 &           $50\times$ &   texture &             Linear Probing &     50.96 \\
    ResNet152 &          1 &           $50\times$ & heightmap &             Linear Probing &     50.96 \\
    ResNet152 &          24 &       $20\times$ + $50\times$ &   texture &               From Scratch &     50.91 \\
     ResNet50 &          1 &       $20\times$ + $50\times$ & heightmap &             Linear Probing &     50.83 \\
         ViTH &          24 &       $20\times$ + $50\times$ & heightmap &             Linear Probing &     50.45 \\
ConvNeXt-large &          24 &           $50\times$ & heightmap &               From Scratch &     50.00 \\
     ResNet50 &          24 &           $50\times$ &   texture &               From Scratch &     50.00 \\
     ResNet50 &          24 &           $50\times$ &   texture & Full-Parameter Fine-Tuning &     50.00 \\
\bottomrule
\caption{All the models and their data configuration that achieve super-human performance (accuracy $>49.5\%$)}
\label{tab:append_superhuman}\\
\end{longtable}}

\end{document}